\documentclass{article}
\usepackage{graphicx}

\begin{document}
\title{The Comparison of Methods for Individual Treatment Effect Detection}
%
%
\author{
	Aleksey Buzmakov\thanks{The publication was prepared within the framework of the Academic Fund Program at the National Research University Higher School of Economics in 2019-2020 (grant No 19-04-048) and by the Russian Academic Excellence Project "5-100".} (abuzmakov@gmail.com) 
	\and Daria Semenova (dvteterina@gmail.com) 
	\and Maria Temirkaeva (mariatemirkaeva@gmail.com) \vspace{5mm} \\
	\it National Research University Higher School of Economics \\
}
\date{}

\maketitle              
\begin{abstract}
Today, treatment effect estimation at the individual level is a vital problem in many areas of science and business. For example, in marketing, estimates of the treatment effect are used to select the most efficient promo-mechanics; in medicine, individual treatment effects are used to determine the optimal dose of medication for each patient and so on. At the same time, the question on choosing the best method, i.e., the method that ensures the smallest predictive error (for instance, RMSE) or the highest total (average) value of the effect, remains open. Accordingly, in this paper we compare the effectiveness of machine learning methods for estimation of individual treatment effects. The comparison is performed on the Criteo Uplift Modeling Dataset. In this paper we show that the combination of the Logistic Regression method and the Difference Score method as well as Uplift Random Forest method provide the best correctness of Individual Treatment Effect prediction on the top 30\% observations of the test dataset.

\providecommand{\keywords}[1]
{
  \small	
  \textbf{\textit{Keywords---}} #1
}\keywords{Individual Treatment Effect  \and ITE \and Machine Learning \and Random Forest \and \texttt{XGBoost} \and \texttt{SVM} \and Random Experiments \and A/B testing \and Uplift Random Forest.}
\end{abstract}
\section*{Introduction}
We live in a world of growing information. Data appear daily in many sectors of our lives. Accordingly, many questions on data storage, processing and usage are arising. In particular, questions about the most suitable methods for certain problems are increasingly being asked. In this paper, we compare several methods for solving one of the most discussed problems of the last decade, individual treatment effect estimation.\par
Under treatments in the academic literature it is customary to understand a certain exposure exerted on an individual or group of individuals in order to provoke a response. So, for example, in marketing push notifications with an appeal to buy goods at a discount may be identified as treatments; in medicine, the definition of treatment is more formal and implies the effect of a drug on the human body. As for the treatment effect, it can be defined as a quantitative indicator that demonstrates how much the treatment has influenced individuals under treatment, compared to individuals from the control group, that is, those who have not been treated.\par
In many situations, individuals may react differently to the same treatments, thus, there is a heterogeneity in responses to a treatment. Given the assumption of a heterogeneous population, we should consider a personalized or individual treatment effect (ITE). There is a number of approaches addressing this problem. So, in our study we will investigate some of the most commonly used approaches for individual treatment effect evaluation, combining them with machine learning methods.\par
Before proceeding to the description of the approaches and methods that we plan to use in our study, it is necessary to introduce several basic  terms related to the topic of ITE evaluation. First of all, we need to identify the outcome or response variable $Y$. This is the parameter that change we need to track during the implementation of a certain treatment. For example, in marketing, such a parameter may be the average customer check after two weeks of an advertising campaign. Usually, outcome variable is numerical or categorical. In the first case, it measures a quantitative indicator, for example, the sum of a check in a particular store, and in the second case, it takes values from some limited set of categories. If only two outcomes are expected, for example, to buy or not to buy something, the variable takes values 1 and 0, respectively, and the outcome is designated as binary. Secondly, it is necessary to identify a set of parameters or characteristics $X$ that affect the target variable. Returning to the previous marketing example, the set of $X$ can consist of recency, frequency and monetary characteristics of particular individual's purchases before the start of the marketing campaign. Thirdly, we should define treatment and control groups. In random experiments, all participants are usually divided into two groups: the treatment group and the control group. The treatment group is the group of individuals that is affected during the experiment, and the control group is the group that is not affected. The results of comparing outcomes in the control and treatment groups allow us to judge the efficiency of the treatment. Now that we have defined the basic concepts, we can move on to the essence of the study.\par
In this paper, we focus on effect for a binary variable and moreover for the approaches of individual treatment effect (ITE) estimation task reduction to classification problem. However our goal is not classification but the identification of the subgroups of clients who are more likely to respond on the treatment positively. We consider several machine learning (hereinafter ML) methods (in particular, linear logit regression, Random Forest, \texttt{\texttt{XGBoost}} and \texttt{SVM}) for three  approaches of treatment effect evaluation: difference score method, modified outcome method and uplift Random Forest~\cite{guelman2015optimal}. To validate the compared approaches on real data, we use Criteo Uplift Modeling Dataset, released along with the paper “A Large Scale Benchmark for Uplift Modeling” by Diemert et al. ~\cite{Diemert2018}.\par
The rest of the paper is organized as follows. The first section reviews the related works. The second section discusses the data we deal with. The third section introduces methodology. The fourth section discusses the evaluation results. The last sections contain conclusions and directions for further research.
\section{Literature review}
To date, there are several review papers that systematize existing methods assessing individual treatment effects (see, for example, ~\cite{jaroszewicz2015ensemble,gutierrez2017causal}. Existing methods, primarily, differ in type of the response or outcome variable $Y$. In most of the studies it is either numerical or categorical. So, if the objective of the study is to evaluate a quantitative indicator, then $Y$ is numerical and corresponds to the numerical value of the outcome variable. For example, in medicine, a numerical variable $Y$ would correspond to the quantity of a certain component in a person blood, and accordingly, treatment will be aimed at a quantitative increase or decrease of this indicator. In other cases the outcome variable is categorical indicating certain states of the individual.In particular, when there are only two possible states, the variable takes the values 0 or 1. In marketing, such a problem can be formulated, for example, as the fact of visiting or purchasing in a certain time window after sending a notification to the customer.\par
Another significant difference between existing models is the methodology underlying them. Here, all models can be divided into two large groups. The first one includes models that transform the task of treatment effect estimation problem into a regression or classification (depending on the type of the target variable) problem in a special way. In particular, let there be some response variable $Y$ and a set of independent variables $X$. The main question in this case is to find the most accurate functional relationship between $Y$ and $X$ $(Y \approx f(X)$). As you can see, there is no treatment variable in the regression and classification problems, and the main issue to be solved in such models is how exactly the treatment variable can be encoded so that the solution of the classification or regression problem gives the solution to the problem of individual treatment effect estimation ~\cite{buzmakov2019machine}. The second large group of methods includes modifications of machine learning techniques specially designed for solving the treatment effect evaluation problem.

\subsection{Reduction to the Classification Task}
Nowadays, methods of reduction to the classification problem are conventionally divided into two groups: indirect estimation methods and direct estimation methods~\cite{guelman2015optimal}. 

Indirect methods are realized through a systematic two-stage procedure of individual treatment effect estimation,the first stage of which is aimed  to achieve high accuracy in predicting the target variable $Y$ due to covariates $X$ and treatment. In the second step, the ITE is estimated as the difference in prediction between the treatment and control group. By contrast, direct methods modify the data in such a way, that only one model is estimated.

\subsubsection{Indirect Methods}
The group of indirect methods includes, for example, the difference-in-difference method~\cite{lo2002true} and the difference score method~\cite{hansotia2002direct,hansotia2002incremental}. In our study we use the difference score method, because ML-methods that we plan to apply are more applicable for it compared to the difference-in-difference method. So let us dwell on difference score method in more detail. 
\paragraph{Difference score method}
The difference score method is also known as the Two-Model approach. The main idea of it lies in the construction of two models that predict the value of the response variable $Y$ by the properties of each description of the object $x \in X$. The first model is built on the treatment group that received the treatment and the second one is built on the control group receiving no treatment. It is assumed that these models can accurately predict the average value of the response variable $Y$ in each of the situations; therefore, the difference in the predictions of the models should give us the average effect for observations with characteristics $X$ ~\cite{buzmakov2019machine}. Because of its simplicity the Two-model approach is usually used as the baseline for more complex statistical and machine learning classification methods~\cite{gutierrez2017causal}, such as logistic regression~\cite{hansotia2002direct,hansotia2002incremental}, Random Forest~\cite{breiman2001random}, \texttt{XGBoost}~\cite{chen2016xgboost} and even neural networks~\cite{manahan2005proportional}.

\subsubsection{Direct methods}
Direct methods represent the idea of adding special interaction variables, which become equal to zero on the control group, and on the test group correspond to independent variables of the original data sample. In this case, a model of the form $\hat{Y}=f(X, T, TX)$ is trained. Such models are well studied in mathematical statistics and econometrics, and, accordingly, there are rigorous methods for calculating the statistical significance and confidence intervals for many specifications of the function $f(\cdot)$, which is an undoubted advantage of this approach~\cite{buzmakov2019machine}. Having trained such a model, the effect of the impact can be estimated as:
\begin{equation}
\bigtriangleup\hat{Y_g}=\hat{\tau}(\delta g)=\hat{\tau}(x)=f(X,1,X)-f(X,0,0 \cdot X)
\end{equation}
The group of direct models includes modified covariate method, causal K-nearest-neighbor, matching before randomization and modified outcome method~\cite{guelman2015optimal}.In our study we will focus on the last of the above approaches due to its ability to be easily combined with machine learning techniques.

\paragraph{Modified outcome method}
The modified outcome method is developed by Jaskowski and Jaroszewicz~\cite{jaskowski2012uplift} and by Weisberg and Pontes~\cite{weisberg2015post} and implies the target variable $Y$ transformation so that the solution of the classification problem on the transformed data provides the solution to the ITE problem. The authors propose setting a new variable $Z$, taking values 1 or 0 in the following way:
\[
    Z_{i} = \left\{
    \begin{array}{ll}
        1 &\textrm{, if  }  T_i=1  \textrm{  and  } Y_i=1\\
        1 &\textrm{, if  }  T_i=0  \textrm{  and  } Y_i=0 \\
        0 &\textrm{,  otherwise}\\
 \end{array} \right. 
 \]
and then fitting a binary regression model to $Z$ on the set of covariates $X$.  
If we accept the assumption that $Y=1$ is more desirable than $Y=0$, we can assume  $W=1$ as an event of receiving a potential treatment result that is at least as good as the observed result~\cite{guelman2015optimal}.\par
In this case, when taking the assumption of the equal size of test and control groups, the expected value of the event $Z = 1$ occurrence will be determined as:
\begin{eqnarray*}
     E[Z]=&E[Y|T=1]\cdot P[T=1] + &\\
    & + E[1-Y|T=0]\cdot P[T=0]=0.5\cdot (1+E[\bigtriangleup Y])
\end{eqnarray*}

Thus, it turns out that the expected value of a new random variable $E[Z]$ monotonously depends on the treatment effect. In this case, the estimated model $Z=f(X)+\epsilon$ will predict conditional expected value of $Z$ on the set of covariates $X$, that is, it will be a monotonous transformation of the treatment effect~\cite{buzmakov2019machine}.\par
The advantages of the modified outcome method seem to be as follow: firstly, it allows to use a wide range of classification models; secondly, it makes possible to get lesser dispersion of model predictions in comparison with the Two-models approach; and thirdly, it is more simple to interpret the coefficients of linear models~\cite{jaskowski2012uplift}.\par
Weisberg and Pontes~\cite{patent:8688610} somehow modified the abovemention approach, entering a random variable $Z = 2 \cdot Y (2 \cdot T - 1)$, expected value of that $E[Z]$ is equal to the expected value of treatment effect $E[T]$. Thus, the model estimation $Z=f(X)+\epsilon $ yields the function $f(X)$, which predicts $E[Z|X]=\tau(X)$. In this case, the entire data sample is used to train the model, and as a result, the quality of prediction of such a model should be better than in the described above model of Jaskowski and Jaroszewicz~\cite{jaskowski2012uplift}. Therefore, by virtue of the assumption of a higher quality model proposed by Weisberg and Pontes~\cite{weisberg2015post}, we will use a modification of these authors in our study. 

\subsection{Uplift random forest models}
In the previous section, we examined in detail the methods related to the group where the task of individual treatment effect estimation is transformed into classification problem. The method described below refers to the second group of approaches mentioned above, involving the use of modified machine learning methods. This method is called Uplift random forest and it was developed by Guelman and his colleagues~\cite{guelman2015uplift}.The main idea of the Uplift Random forest is the tree-building algorithm with the sensitive splitting criteria, proposed by Rzepakowski and Jaroszewicz~\cite{rzepakowski2012decision}. The method is based on the criteria for splitting tree nodes, which in turn is intended to maximize the distance between the empirical distribution density of the response variable Y in the test and control groups. As a distance, the authors consider the Kullback-Leibler divergence. The quality of a particular splitting is characterized by growth rate of the distance between test and control group. In this work we consider for a reference only one method from this group since the main  idea of this paper is to check whether machine learning models can improve the estimation quality of reduction approaches.
\section{Data}
To compare different methods for treatment effect evaluation, we used Criteo Uplift Modeling Dataset. The choice of the dataset was justified by its large dimension, as well as the  data openness. This dataset was created by the Diemert and his colleagues~\cite{Diemert2018} by combining the data obtained as a result of several incrementality tests, a specific randomized research procedure, in which advertising treatment directed not to all customers, but only to a randomly selected test group. The dataset consists of 25 million observations, each of which represents a user of the website. For each person it is known his 12 features, belonging to a group (test or control), whether the user visited and/or converted on the advertiser's website during the test period (2 weeks), and whether the user has been effectively treated. Due to the loss of potential income in the control group, advertisers usually keep only a small control group.\par 
Criteo Uplift Modeling Dataset contains significantly modified data by the Diemert and his colleagues for the reason of privacy~\cite{Diemert2018}. The initial data was processed as follows: the data was sub-sampled non-uniformly so the initial level of increment cannot be deduced; the names of the features were anonymous, and their values were randomly designed to preserve predictive power, while it is almost impossible to restore the original features or user context.
\section{Methodology}
The main goal of this study is the comparison of different machine learning methods applied to different approaches of individual treatment effect evaluation. We aimed to compare various methods in terms of better identification of clients who are more likely to respond to treatment, i.e. who have a higher individual treatment effect (ITE):
\begin{equation}
    ITE=E[Y_i=1\mid X_i, treatment=1]-E[Y_i=1\mid X_i, treatment=0]
\end{equation}

Each client was assigned either to a test group, that is, to the one that received the marketing treatment, or to the control group, the participants of which did not receive marketing treatment. Note that in this case it is impossible to observe the same person in two groups at the same time, which makes the task of ITE estimation for an individual point infeasible. But despite this, there are several approaches to solving it, some of which have been described in the literature review - so, we will use them in our study.\par
We apply linear logistic regression, Random Forest, \texttt{XGBoost} and \texttt{SVM} methods to Difference Score and Modified Outcome approaches of measurement of ITE, as well as Random forest method to Uplift modeling, proposed by Guelman~\cite{guelman2015optimal}. The choice of above mentioned machine learning methods was justified by their popularity and frequency of use in studies of the treatment effect evaluation. Presumably, the Modified Outcome method and Uplift modeling should show better results compared to the indirect method, since they are aimed directly at ITE modeling, in contrast to the latter approach.\par
So, the main steps of the experimental part of the study are described below.
\begin{description}
\item[Preliminary data analysis.]
According to a preliminary analysis of the data, the response of customers in the test and control groups was 4.41\% and 2.61\%, respectively. Results of t-test confirm the statistically significant difference of target variable between the test and control groups, which, in turn, indicates the presence of the treatment effect (p-value < 0.01), but this difference is small. That is why it is extremely important to find a combination of approach and method which will allow to estimate the individual treatment effect most accurately. Our algorithm, described below, is aimed at achieving this goal.

\item[Random split.] We randomly assigned clients to either training (sample used for learning model) or holdout group (sample used for test of evaluation quality). The size of treatment and control group in training dataset are equal and less than in original dataset. We divide entire sample into training and holdout sets because training and evaluate quality of model on the same sample leads to retraining (to find patterns in the training sample that are not patterns for all other data). Only test of evaluation quality on holdout set will correctly convey the quality of the model.
\item[Model estimation.] Fit the models on training sample. In reason  of the computational complexity the training sample for \texttt{SVM} and Uplift Random forest models was truncated.
\item [Prediction.] Predict ITE on holdout sample. These values of ITE are ordered from high to low and binned together in deciles. We limit quantity of targets clients by 30\% with the highest score of ITE similarly Guelman~\cite{guelman2015uplift}. This action can be explained by two reasons. First of all, all individual considered by model as being persuadable will have a high ITE and thus be in front. Secondly, there is also practical reason: the cost of marketing campaigns is usually high, therefore, firms try to limit the share of targets clients.
\item [Quality metrics of models.] Average treatment effect and total treatment effect are used to compare different methods and choose the best. The ideal model will predict large ITE for observations with real large ITE, and as a result, the average effect should be large in the selected subsample of the holdout sample. On the other hand, if the model gives poor predictions and, as a consequence, arranges the observations in a random order, then the average treatment effect will not differ from the average effect on the entire holdout sample. This means that different models for assessing the ITE can be compared by the size of the AVE on subsample. In most marketing tasks, the researcher is not so much interested in the ATE as the overall effect (TTE). TTE is ATE multiplied by the number of target customers.
\end{description}

\begin{equation}
ATE=E[Y\mid treatment=1]-E[Y\mid treatment=0]
\end{equation}
\begin{equation}
TTE=ATE\cdot N
\end{equation}
where $N$ is the size of subgroup.\\

Another vital step that must be taken after calculating the quality metrics is to verify the performance of the predictive model in practice. To do this, we use the cross-validation method. Cross-validation iterations are performed for different divided sets, which helps to reduce the spread of results, and the verification results are averaged over all iterations. In our study, each cross-validation iteration includes all algorithm steps from random split to model quality metrics measure. In each iteration new training and holdout samples are submitted for input. Then iterations are performed twenty times to reduce the results’ spread whereupon results are averaged over all iterations. 
\section{Empirical results}
In this section we  demonstrate the results of different ML-methods implementation to three approaches of ITE detection. It should be recalled that we studied the interaction of such approaches as modified outcome method (mom) and difference score method (2M)  with the following ML-methods: Logit, Random Forest, \texttt{SVM} and \texttt{XGBoost}. Also, we compared all the above combinations with a special model - Uplift Random Forest. All approaches were implemented using the twenty-fold cross-validation procedure. Table 1 allows comparing the performance of different ML methods applied to three approaches of ITE detection in terms of average treatment effect of subgroup, particularly, of top 30\% clients with the highest ITE.
\begin{table}[t]
\centering
\label{tab1}
\caption{Performance of different approaches and ML methods}
\begin{tabular*}{\textwidth}{@{\extracolsep{\fill}}c|cc|cc}
\hline
 &  \multicolumn{2}{c}{TTE}   & \multicolumn{2}{c}{ATE} \\
\multicolumn{5}{l}{Difference score method}  \\
\hline
 & Mean &  Conf. Int.   &  Mean & Conf. Int. \\
\hline
Logit &  6970.98  & 6856.51-7085.44 &  0.05  & 0.048-0.05 \\
Random Forest &  6075.33  & 5955.31-6195.36 &  0.014  & 0.0138-0.0143 \\
\texttt{XGBoost} &  5448.48  & 5328.58-5568.37 &  0.039  & 0.038-0.039 \\
\texttt{SVM} &  2996.21  & 1829.73-4162.70 &  0.020  & 0.012-0.028 \\
\hline
\multicolumn{5}{l}{Modified outcome method}  \\
\hline
 & Mean &  Conf. Int.   &   Mean & Conf. Int. \\
\hline
Logit & 1676.63 & 1616.65-1736.61 & 0.01 & 0.011-0.012 \\
Random Forest & 126.67 & 115.26-138.09 & 0.0009 & 0.0008-0.0009 \\
\texttt{XGBoost} & 2898.38 & 2825.84-2970.93 & 0.02 & 0.021-0.021 \\
\texttt{SVM} & 501.99 & 120.44-883.55 & 0.004 & 0.001-0.006 \\
\hline
\multicolumn{5}{l}{Uplift}  \\
\hline
 & Mean &  Conf. Int.   &   Mean & Conf. Int. \\
\hline
Random Forest & 6145.43 & 5938.28-6352.57 & 0.04 & 0.039-0.042 \\
\hline
\end{tabular*}
\end{table}

It is worth noting, that all methods give different estimates with different confidence intervals.\par
According to the results obtained, difference score method estimated by linear logistic regression turned out to be the best among all. Also, we should highlight the significant statistical difference in mean values ATE and TTE between two best models: 2M-GML and UPLIFT-RF. Therefore, despite the fact that the difference between the mean values of ATE and TTE of two best models is not so large, mean values differ statistically, which means that 2M-GML model is better than UPLIFT-RF model in terms of ATE and TTE.\par
One of the reason for the high quality of the Two-model approach could be the particularity of the dataset. In particular, if the model $\hat{Y}=f(X,\textrm{treatment})$ is well fitted in the data, then the two model can provide good results. Thus method comparison on other datasets is needed to check this statement.\par
\begin{figure}[t]
\includegraphics[width=\textwidth]{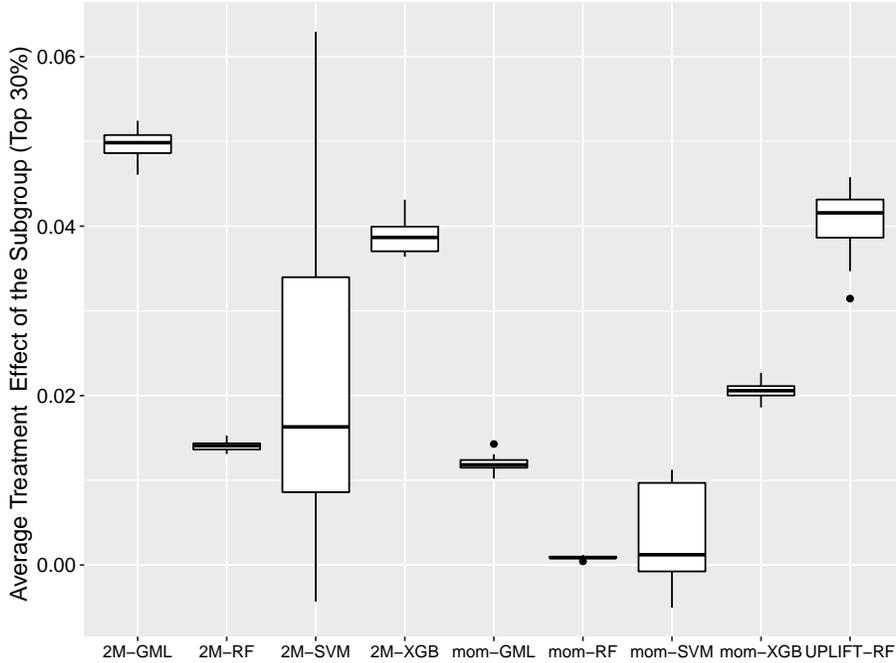}
\caption{Boxplots of the Average Treatment Effect measure for all models. The dots outside the boxplots represent outliers. We used “1.5 rule” for determining if a data point is an outlier: less than Q1-1.5*(Q3-Q1) or greater than Q3+1.5*(Q3-Q1), where Q1 and Q3 represent the first and third quartiles, respectively.} \label{Fig1}
\end{figure}

Another important point to be made relates to computational complexity of Uplift Random Forest the train dataset for this method was truncated. Probably, this fact has become the reason for the poorer performance  demonstrated by UPLIFT-RF compared to 2M-GML. With regard to the worst results, they were demonstrated by combinations of the \texttt{SVM}-method and modified outcome approach, as well as of the Random forest and modified outcome approach. A visual representation of the ATE results, obtaining due to the twenty-cycles cross-validation is presented on the Figure \ref{Fig1}.

\section{Conclusion}
Nowadays,  heterogeneous treatment effect estimation is a fundamental and useful for business and science problem. There are various approaches have been proposed to identify the individuals that are most likely to respond towards a treatment. In this paper we were aimed to compare some of the well-known approaches such as Modified Outcome method, Difference Score method and Uplift Random Forest. The first two approaches of ITE measurement  were estimated by linear logistic regression, Random Forest, \texttt{SVM} and \texttt{XGBoost}. The Criteo Uplift Modeling Dataset was used for fitting models.\par
In this study, we used two metrics: average treatment effect (ATE) and total treatment effect (TTE) on three deciles predicted by the model. To reduce the noise rate of results cross-validation procedure was performed on different splitted subsamples. \par
According to the empirical results, Difference Score method modeling by linear logistic regression and Uplift Random Forest have achieved the best performance. The noisiest results were showed by \texttt{SVM} in combination with both Modified Outcome and Difference Score approaches. Since the conclusion on the linear regression best perfomance seems to be really surprising, proposed methodology needs to be further verified with other data.\par
The practical implementation of the research results will presumably consist in applying the best combination of approach and method of ITE estimation to real marketing campaigns, which, in turn, will increase their effectiveness in terms of consumers' responses rise.

\section{Future research}
Finally, there are some limitations of paper that open area for further research. We consider the case of reduction to classification problem, reduction to regression problem also may be useful in terms of financial gain from a marketing campaign. Also we have used only three approaches of individual treatment effect estimation. It is available to extent list of approaches by, for example, modified covariate method~\cite{tian2014simple}. Moreover, because of the computation complexity a small share of the initial database was used for training set, increasing the size of the training set can improve the models quality.

\bibliographystyle{splncs04}
\bibliography{bibliography.bib}

\end{document}